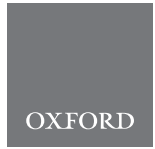

# DeepDiff: Deep-learning for predicting Differential gene expression from histone modifications

**Arshdeep Sekhon, Ritambhara Singh, and Yanjun Qi** *

Department of Computer Science, University of Virginia, Charlottesville, VA, U.S.A

## Abstract

**Motivation:** Computational methods that predict differential gene expression from histone modification signals are highly desirable for understanding how histone modifications control the functional heterogeneity of cells through influencing differential gene regulation. Recent studies either failed to capture combinatorial effects on differential prediction or primarily only focused on cell type-specific analysis. In this paper we develop a novel attention-based deep learning architecture, DeepDiff, that provides a unified and end-to-end solution to model and to interpret how dependencies among histone modifications control the differential patterns of gene regulation. DeepDiff uses a hierarchy of multiple Long short-term memory (LSTM) modules to encode the spatial structure of input signals and to model how various histone modifications cooperate automatically. We introduce and train two levels of attention jointly with the target prediction, enabling DeepDiff to attend differentially to relevant modifications and to locate important genome positions for each modification. Additionally, DeepDiff introduces a novel deep-learning based multi-task formulation to use the cell-type-specific gene expression predictions as auxiliary tasks, encouraging richer feature embeddings in our primary task of differential expression prediction.

**Results:** Using data from Roadmap Epigenomics Project (REMC) for ten different pairs of cell types, we show that DeepDiff significantly outperforms the state-of-the-art baselines for differential gene expression prediction. The learned attention weights are validated by observations from previous studies about how epigenetic mechanisms connect to differential gene expression.

**Availability:** Codes and results are available at deepchrome.org

**Contact:** yanjun@virginia.edu

## 1 Introduction

Gene regulation is the process of controlling gene expression. The human body contains hundreds of different cell types. Although these cells include the same set of DNA information, their functions are different. Cells resort to a host of mechanisms to regulate genes differently. Many factors, especially those in the epigenome, can affect how cells express genes differently. As reviewed in [11, 25], epigenomics studies how gene expression is altered by a set of chemical reactions over the chromatin that do not alter the DNA sequence.

Histone modification(HM) is one set of critical chemical reactions over the chromatin that plays a crucial role in regulating gene transcription. DNA strings are wrapped around 'bead'-like structures called nucleosomes that are composed of histone proteins. These histone proteins are prone to a variety of modifications (e.g., methylation, acetylation, phosphorylation, etc.) that can modify the spatial orientation of the DNA structure. Such modifications impact the binding behavior of transcription factor proteins (to DNA) and thus generate different forms of gene regulation. The significant role of histone modifications in influencing gene regulation was evidenced in studies like connecting anomalous histone modification profiles to cancer occurrences [3]. Contrary to DNA mutations, such epigenetic changes are potentially reversible ([3]). This vital feature has brought histone modifications to the center stage of epigenetic therapy.

Recent advances in next-generation sequencing have allowed researchers to measure gene expression and genome-wide histone modification patterns as read counts across many cell types. These datasets have been made available through large-scale repositories, one latest being the Roadmap Epigenome Project (REMC, publicly available) ([22]). REMC has released thousands of genome-wide datasets including gene expression reads (RNA-Seq datasets), and HM reads across 100 different human cells/tissues [22]. Multiple recent papers tried to understand gene regulation by predicting gene expression from large-scale HM signals. Related studies (summarized in Appendix Table 2) have mostly focused on the formulation under a single cell condition, even though gene regulation undergoes differential changes by environmental triggers, from one tissue type to another, or under different cell development stages.

Differential gene expression, or difference in expression levels of the same gene in two cell conditions, controls functional and structural heterogeneity of cells and has also been implicated in a number of diseases, providing valuable tools for the discovery of therapeutic targets and diagnostic markers. Differential gene expression has been linked to aberrant HM profiles in the literature. For example, [13] showed the correlation between differential gene expression in different stages of Alzheimer's disease-like neurodegeneration in mice. Further, the authors observed that the changes in HM patterns associate with the differentially regulated genes. [20] reported coordinated changes between HM profiles and differential gene expression across the lymphoblastoid cell line GM06990, K562, and HeLa-S3 cell lines. As another example, [32] showed links between differential expression of naive T cells vs. memory T cells, ascribed mainly to changes in histone modifications.

This paper proposes an attention-based deep learning architecture to learn from datasets like REMC, how different histone modifications work together to influence genes' differential expression pattern between two different cell-types. We argue that such differential analysis and differential understanding of gene regulation from HMs can enable new insights into principles of life and diseases, will allow for the interrogation of previously unexplored regulation spaces, and will become an important mode of epigenomics analysis in the future. Four fundamental data challenges exist when modeling such tasks through machine-learning:

1. Genome-wide HM signals are spatially structured and may have long-range dependency. For instance, to quantify the influence of a histone modification mark, learning methods typically need to use as input features all of the signals covering a DNA region of length $10,000$ base pair (bp) centered at the transcription start site (TSS) of each gene. These signals are sequentially ordered along the genome direction. To develop "epigenetic" drugs, it is important to recognize how an HM mark's influence varies over different genomic locations.

2. The core aim is to understand what the relevant HM factors are and how they work together to control differential expression. Various types of HM marks exist in human chromatin that can influence gene regulation. For example, each of the five standard histone proteins can be simultaneously modified with various kinds of chemical modifications, resulting in a large number of varying histone modification marks. As shown in Figure 1, we build a feature vector representing signals of each HM mark surrounding a gene's TSS position. When modeling genome-wide signal reads from multiple marks, learning algorithms should take into account the modular nature of such feature inputs, where each mark functions as a module. We want to understand how the interactions among these modules influence the prediction (differential gene expression).

3. Since the fundamental goal of such analysis is to understand how HMs affect gene regulation, it requires the modeling techniques to provide a degree of interpretability and allowing for automatically discovering what features are essential for predictions.

4. There exist a small number of genes exhibiting a significant change of gene expression (differential patterns) across two human cell types like A and B. This makes the prediction task using differential gene expression as outputs much harder than predicting gene expression directly in a single condition like A alone or B alone.

In this paper, we propose an attention-based deep learning model, DeepDiff, that learns to predict the log-fold change of a gene's expression across two different cell conditions (assuming cell type A and cell type B in the rest of the paper). Here, $\boldsymbol{X}^A \in \mathbb{R}^{M \times T}$ and $\boldsymbol{X}^B \in \mathbb{R}^{M \times T}$, where $M$ represents the number of HM signals and $T$ represents the number of bins. For each gene, its input signals consist of $\boldsymbol{X}^A$ (the histone modification signals from A), $\boldsymbol{X}^B$ (the histone modification signals from B) and $(\boldsymbol{X}^A - \boldsymbol{X}^B)$ (the difference matrix between HM signals of these two conditions). All three cover the gene's neighboring $10,000$ base pair regions centered at the transcription start site (TSS). (1) To tackle the first challenge of modeling spatially structured input signals, we use the Long Short-term Memory (LSTM) (Section 3.4) deep-learning module that can represent interactions among signals at the different positions of a chromatin mark. Because we model multiple HM marks, resulting in multiple LSTMs learning to embed various HM marks. (2) To handle the second challenge of modeling how HM marks work together, we use a second-level LSTM to learn complex dependencies among different marks. (3) For the third challenge of interpretability, we borrow ideas from the AttentiveChrome[29] that focuses on cell-specific predictions. We train two levels of "soft" attention weights, to attend to the most relevant regions of a chromatin mark, and to recognize and attend to the critical marks for each differential expression prediction. Through predicting and attending in one unified architecture, DeepDiff allows users to understand how chromatin marks control differential gene regulation between two cell types. (4) For the last challenge of difficult label situation for differential expression prediction, we design a novel multi-task framework to use the cell-type-specific prediction network as auxiliary tasks to regularize our primary task of differential expression prediction. The cell-type specific system (one for cell A and another one for cell B) also uses attention plus the hierarchical LSTMs formulation. Further, we introduce an additional auxiliary loss term that encourages the learned embeddings of HM inputs $\boldsymbol{X}^A$ and $\boldsymbol{X}^B$ to be far apart for differentially expressed genes.

In summary, DeepDiff provides the following technical contributions:

- DeepDiff uses a hierarchy of two levels of gene-specific attention mechanisms to identify salient features at both the bin level and the HM levels. It can model highly modular inputs where each module is highly structured. Attention weights enable our model to explain its decisions naturally by providing "what" and "where" in HM signal inputs is important for the differential gene expression output. This flexibility and interpretability make this model an ideal technique for handling large-scale epigenomic data analysis.

- We introduce an auxiliary task and auxiliary loss formulation to aid the main task of differential gene expression prediction. The proposed multitasking framework couples the two related tasks of cell-type-specific predictions and the main task of differential expression prediction. This auxiliary formulation provides the model with additional information from the auxiliary evidence, encouraging richer feature embeddings compared to only difference HM features. It helps DeepDiff to build on top of the state-of-the-art AttentiveChrome[29] and can borrow auxiliary features from AttentiveChrome tasks. Further, we introduce a novel auxiliary loss term inspired by the contrastive loss[15] of the Siamese architecture formulation. This loss term encourages the model to learn embeddings whose neighborhood structures in the model's representation space are more consistent with the differential gene expression pattern.

- To the authors' best knowledge, DeepDiff is the first deep learning based architecture for relating histone modification and differential gene expression patterns. DeepDiff provides more accurate predictions than state-of-the-art baselines. Using datasets from REMC, we evaluate DeepDiff on ten different pairs of cell types. We validate the learned attention weights using previous observations obtained from HM enrichment analysis across differentially regulated genes.

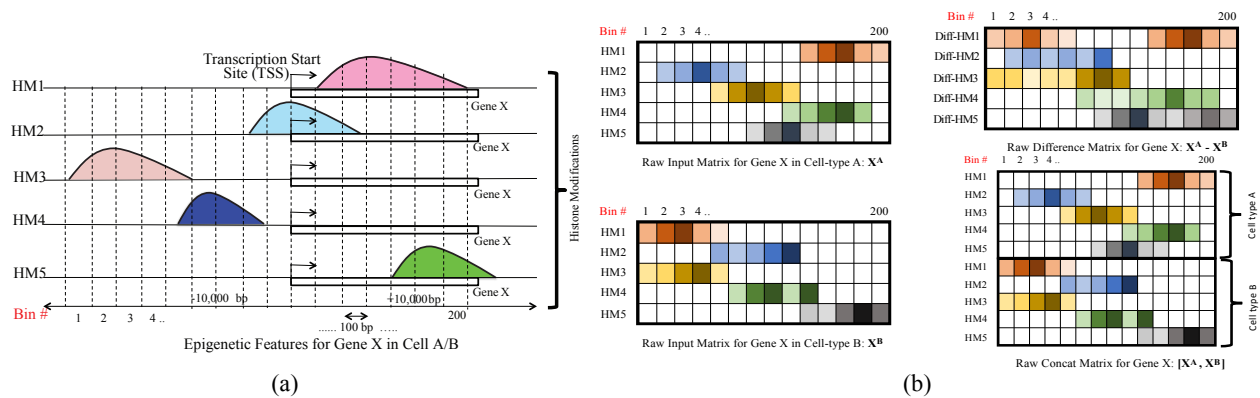

**Fig. 1.** (a) Feature input generation for a gene in a cell-type and (b) Raw input feature variations to DeepDiff model for differential expression: difference and concatenated HM signals of both cell-types.

## 2 Previous Works

Multiple computational methods have been proposed to employ HMs for predicting gene expression using large-scale histone modification datasets. Recent methods in the literature can be roughly grouped into three categories with respect to the formulation of outputs: regression, classification or ranking. (1) The regression-based models include linear regression[18, 9] and Support Vector Regression (SVR)[6]. [6] divided the DNA regions around TSS (transcription start site) and TTS (transcription terminal site) into small bins of 100 base pairs and used a multiple bin-specific Support Vector Regression (SVR) to model HM signals for gene expression prediction. They extended this SVR model to predict differential gene expression between mouse embryonic stem cell and neural progenitor cell using the difference of the HM signals as features/inputs. This study also uses a two-layer SVR model to integrate information from multiple bins. The first layer is a bin-specific SVR model for histone modification features. A second layer takes as input the predictions from the first layer across all bin positions and predicts a single regression output for differential gene expression. (2) [12] proposed a ranking based cell type-specific model that formulates gene expression prediction as a ranking task such that high ranks correspond to high levels of gene expression and low rank corresponds to low expression. (3) Multiple studies used classification-based formulation to model the gene expression prediction from HM inputs. This includes support vector machines[6], random forests[10, 23], rule-based learning[16] and deep learning frameworks like DeepChrome[28] and AttentiveChrome[29]. [10] used a random forest classifier to predict genes as silent or transcribed, while the authors also used linear and multivariate regression to predict gene expression values. [23] presented a two-step process- feature selection, then followed by prediction for differential gene expression. It used the so-called ReliefF[21] based feature selection and Random Forest Classification. DeepChrome[28] and Attentive Chrome[29] are deep learning based frameworks for cell-type specific gene expression prediction. DeepChrome[28] used Convolution Neural Nets (CNN). Differently, AttentiveChrome[29] used a hierarchical attention-based deep learning architecture to predict gene expression from HM reads. Appendix Table 2 compares all the aforementioned related studies for gene expression prediction.

## 3 Method

### 3.1 Background: Recurrent Neural Networks and Long Short-Term Memory (LSTM)

Recurrent neural networks (RNNs) have achieved remarkable success in sequential modeling applications like translation, image captioning, video segmentation, etc. A sequential input of RNN is normally represented by an input matrix $\mathbf{X}$ of size $n_{in} \times T$, where $T$ represents the time steps and $n_{in}$ represents the dimension of the features describing each timestep of the input. For an input $X$, an RNN produces a matrix $\mathbf{H}$ of size $D \times T$ as output, where $D$ is the RNN embedding size. More concretely, at each timestep $t \in \{1, \cdots, T\}$, an RNN takes an input column vector $\mathbf{x}_t \in \mathbb{R}^{n_{in}}$ and the previous hidden state vector $\mathbf{h_{t-1}} \in \mathbb{R}^d$ to produce the next hidden state $\mathbf{h_t}$ by applying the following recursive operation:

$$\mathbf{h}_t = \sigma(\mathbf{W}\mathbf{x}_t + \mathbf{U}\mathbf{h}_{t-1} + \mathbf{b}) = \overrightarrow{LSTM}(\mathbf{x}_t), \quad (1)$$

where $\mathbf{W}, \mathbf{U}, \mathbf{b}$ are the trainable parameters of the model, and $\sigma$ is an element-wise nonlinearity function. Due to the recursive nature, in theory, RNNs can capture the complete set of dependencies among all time steps without having to learn different parameters for each time step, like all spatial positions in a sequential sample.

A variant of the RNN, LSTM[17], further improves upon the basic RNN(Eq. (1)) to model long-term dependencies. In addition to the hidden state-to-state recurrent component in an RNN, an LSTM layer has a recurrent cell state updating function and gating functions. The gating functions control what information needs to be added or removed from the cell state. This combination of cell state and gating functions allows the LSTM to learn long-term dependencies while avoiding vanishing and exploding gradients. Similar to a basic RNN, when given input vector $\mathbf{x}_t$ and the state $\mathbf{h}_{t-1}$ from previous time step $t-1$, an LSTM module also produces a new state vector $\mathbf{h}_t$. For our task, we call each bin position on the genome coordinate a "time step".

### 3.2 Attention-based deep-learning models

Deep neural networks augmented with attention mechanisms have obtained great success on multiple artificial intelligence topics such as machine translation ([2]), object recognition ([1, 26]), image caption generation ([34]), question answering ([30]), text document classification ([35]), video description generation[36], visual question answering[33], or solving discrete optimization [31]. The idea of attention in deep learning is inspired by the properties of the human visual system. When perceiving a scene, the human vision fixates more on some areas over others, depending on the task at hand ([8]). Augmenting deep learning models with attention allows them to focus selectively on only relevant features for a prediction. Different attention mechanisms have been proposed in the literature, including 'soft' attention [2], 'hard attention' [34, 24], or 'location-aware attention' [7]. Soft attention [2] calculates a 'soft' weighting scheme over all the components of an input. These weights indicate the relative importance of each feature component for a given prediction. The weights are then used to compute a summary representation of the input as a weighted combination of the components. The magnitude of an attention weight correlates highly with the degree of significance of the corresponding component to the prediction. This property is particularly ideal for adapting deep learning to biology tasks, as it gives users interpretable information regarding how features contribute to a prediction. Recently, AttentiveChrome[29] introduced two levels of attention at the bin and the histone-modification level, enabling the user to get information about which features were responsible for each prediction at the sample level.

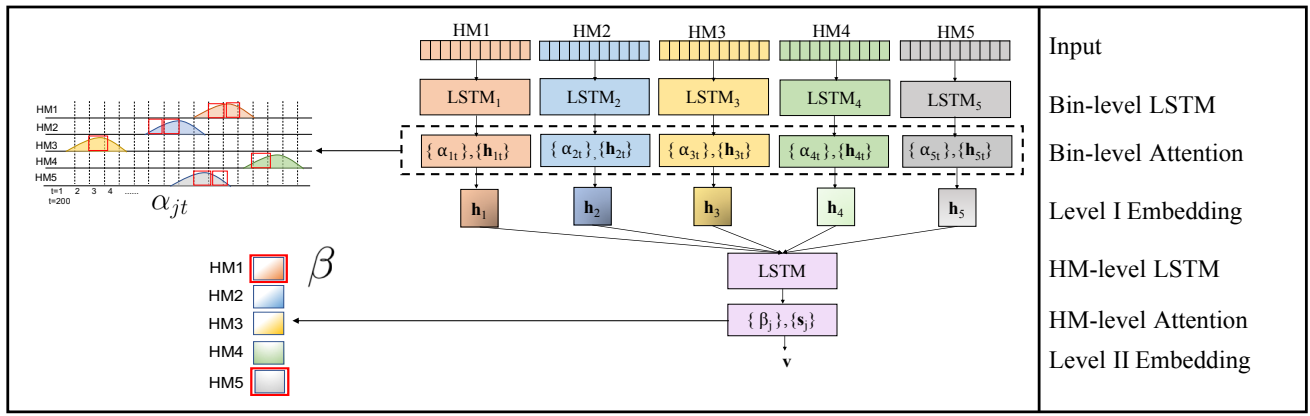

**Fig. 2.** Two level attention mechanism used in DeepDiff variations for meaningful feature representation: $\alpha_{jt}$ represents the bin level attention for HM $j$ and bin $t$ obtained from the Level I Embedding module attention mechanism, indicating the relative importance for bin $t$ in HM $j$. $\beta_j$ represents the HM-level attention for HM $j$ obtained from the Level II Embedding module's attention mechanism, representing the relative importance of HM $j$.

### 3.3 Input Generation

We focus on the predictive modeling of differential gene expression given the histone modification profiles of a gene in two cell-types. We formulate the output as the log fold change in expression given the histone modification profiles for the two cell-types under consideration. Similar to DeepChrome[28] and AttentiveChrome[29], we divided the 20, 000 basepair (bp) DNA region (+/− 10000 bp) around the transcription start site (TSS) of each gene into bins of length 100 bp. Each bin includes 100 bp long adjacent positions flanking the TSS of a gene. We consider five core histone modification marks that have been uniformly profiled across multiple cell types in the REMC database ([22]). Appendix Table 4 summarizes the 5 HMs we use and their associated functional regions on the genome. Figure 1(a) summarizes our input matrix generation strategy. More concretely, our input for each gene includes two 5 × 200 matrices, each matrix corresponding to each of the two cell types under consideration(depicted in Figure 1(b)). Columns and rows in each matrix represent bins and histone modifications, respectively. Thanks to the capability of neural networks for learning meaningful representations, we do not perform any feature selection before feeding the matrices in Figure 1(b) to the proposed DeepDiff model variations, a hierarchical attention-based DNN. Compared to previous studies (Appendix Table 2), DeepDiff does not need to explore the best-bin, averaged or other feature selection strategies. As an end-to-end strategy(raw features to predictions), DeepDiff eliminates the need to evaluate different feature selection strategies.

### 3.4 DeepDiff : Learning meaningful representations through two levels of embedding and attention modules

*Notations:* Our training set consists of $N_{samp}$ gene samples in the form of $(\mathbf{X}_{(n)}, y_{(n)})$ pairs, where $n \in \{1 \cdots N_{samp}\}$. Given two cell-types $A$ and $B$, and a gene $g$ under consideration, the HM profile of gene $g$ in $A$ and $B$ is denoted as $\mathbf{X}^A$ and $\mathbf{X}^B$, respectively. We consider $M = 5$ HM marks for each gene. Each HM signal across the $T = 200$ bins in cell-type $A$ is represented by a row vector in $\mathbf{X}^A$. Similarly, for cell-type $B$, each HM signal is represented by a row vector in $\mathbf{X}^B$.

We do not perform any feature selection on the raw histone modification features. Instead, we use deep learning modules to learn sensible features. Our raw features have two important properties (as depicted in Figure 1):(1) In addition to the raw HM signals from the two cell types under consideration, we use difference and concatenation of the raw HM signals. This results in modular raw input features with four possible input matrices, rows corresponding to HM vectors (as shown in Figure 1(b)). (2) These HM vectors are spatially structured along the genome coordinate. Considering the spatially structured raw features and their modular property, we use two levels of basic embedding modules: Level I and Level II embedding units coupled with two levels of attention modules. These basic modules used in DeepDiff variations are illustrated in Figure 2. Now, we explain how we use deep learning to learn the representation of each matrix in Figure 1(b).

*Level I Embedding ($f_1$):* The Level I Embedding module consists of a bin-level LSTM for learning the embedding of every HM, followed by a bin level attention mechanism. The bin level LSTM sequentially models the signal at each bin position. We name LSTMs, for all input HMs, put together, as the Level I Embedding module. The Level I Embedding module $f_1$ consists of multiple bidirectional LSTMs, one for each input HM. The $LSTM_j$ corresponding to HM $j$, takes as input the $jth$ HM,i.e, $jth$ row vector in matrix $\mathbf{X}$, $\mathbf{X}_j = [\mathbf{x}_{j1}, \ldots, \mathbf{x}_{jt}, \ldots, \mathbf{x}_{jT}]$. A bidirectional LSTM has one LSTM in each direction. The forward LSTM models dependencies in $\mathbf{x}_j$ in the direction 1 to T, i.e., $\overrightarrow{\mathbf{h}}_{jt} = \overrightarrow{LSTM}_j(\mathbf{x}_{jt})$ where $\overrightarrow{\mathbf{h}}_{jt}$ is of hidden state size $D$. The backward LSTM models the dependencies from $T$ to 1: that is $[\mathbf{x}_{jT}, \ldots, \mathbf{x}_{jt}, \ldots, \mathbf{x}_{j1}]$. Hence, $\overleftarrow{\mathbf{h}}_{jt} = \overleftarrow{LSTM}_j(\mathbf{x}_{jt})$. The output of the bidirectional LSTM is a concatenation of the hidden state output of the forward and backward LSTMs at each $t$ position: $\boldsymbol{h}_{jt} = [\overrightarrow{\boldsymbol{h}}_{jt}, \overleftarrow{\boldsymbol{h}}_{jt}]$, where $\boldsymbol{h}_{jt}$ is of size $2 \times D$.

*Bin level Attention* : Attention in Deep Learning is a powerful tool used to highlight features that are important for a given prediction. The hidden state at each step of the LSTM produces an embedding for that bin position $\boldsymbol{h}_{jt}$. To get a cumulative embedding to represent all the bin positions, one strategy could be to sum the embeddings across all bin positions. However, not all the bin positions are equally relevant. For example, in certain HM patterns, bin positions near the transcription start site are more important than the ones away from it. To learn and encode such important information into the embeddings, we use a soft attention mechanism that automatically discovers which are the important bin positions as part of the training process. The attention-augmented LSTMs result in learning a weighting representation for the bin embeddings, such that more important bin positions get a higher weight. In detail, this is done using a context weight vector, denoted by $\boldsymbol{Wb}_j$ of dimension $2 \times D$ for each HM $j$. An attention weight $\alpha_{jt}$, corresponding to bin position $t$ for the $jth$ HM is obtained by

$$\alpha_{jt} = \frac{exp(\boldsymbol{h}_{jt} \cdot \boldsymbol{Wb}_j)}{\Sigma_{k=1}^{T}(exp(\boldsymbol{h}_{jk} \cdot \boldsymbol{Wb}_k))} \quad (2)$$

where $\boldsymbol{Wb}_j$ is learned through training and $\cdot$ indicates dot product. These attention weights are then used to weigh the embedding vectors of all bins to get a summary embedding: $\boldsymbol{h}_j = \Sigma_{t=1}^{T}(\alpha_{jt} \times \boldsymbol{h}_{jt})$. Essentially, this "summary" representation for each HM represents a bin importance weighted sum of all bins in the HM under consideration. The attention tells us where in this HM is important for prediction.

*Level II Embedding ($f_2$):* To efficiently represent the combinatorial dependencies between the various HMs, we use another LSTM as a second level embedding module. This LSTM takes as input the Level I Embedding output $\boldsymbol{h}_j$ where $j \in \{1, \ldots, M\}$. In detail, the $jth$ HM embedding from Level I Embedding module $\boldsymbol{h}_j$ is used as input to the $jth$ time step in a bidirectional LSTM. The LSTM will generate an embedding vector for $jth$ time step: $s_j = LSTM(\boldsymbol{h}_j)$.

*HM level Attention* : To combine the outputs from all $M$ HMs in an informative way, we use a second level attention mechanism to learn attention weights $\beta_j$, representing the importance of the $jth$ HM. To get these HM level attention scores $\beta_j$ where $\{j \in 1, \ldots, M\}$, we learn an HM-level context vector $\boldsymbol{W}_h$ to calculate an attention score as

$$\beta_j = \frac{exp(\boldsymbol{s}_j \cdot \boldsymbol{W}_h)}{\Sigma_{l=1}^{M}(exp(\boldsymbol{s}_l \cdot \boldsymbol{W}_h))} \quad (3)$$

This attention weight $\beta_j$ intuitively represents the relative contribution of the HM $\boldsymbol{x}_j$ to the summary representation of the whole matrix $\boldsymbol{X}$. To get a summarized embedding of the HMs, we use the outputs at all time steps of the HM level LSTM weighted by its attention score as the final embedding of describing $\boldsymbol{X}$ i.e. $\boldsymbol{v} = \Sigma_{j=1}^{M}(\beta_j \times \boldsymbol{s}_j)$. Including the bin level as well as HM-level attention weights representing $X$ allows us to interpret which bins in which HMs were relatively more important for the current prediction.

### 3.5 DeepDiff Main Task: an End-to-End Deep-Learning Architecture for Regression

We formulate differential gene expression prediction (our main task) as a regression task. The target label for a gene is the log fold change of its expression between the two cell-types under consideration. The learned representation vector **v** from Level II Embedding is fed into a multi-layer perceptron (MLP) module to learn a regression function, a mapping from HM profiles to the target real value representing differential gene expression. In detail, this prediction module $f_{mlp}(.)$ comprises a standard, fully connected multi-layer perceptron network with multiple alternating linear and non-linear layers. Each layer learns to map its input to a hidden feature space, and the last output layer learns the mapping from the hidden space to the output label space. The whole network output can be written as:

$$f(\mathbf{X}) = f_{mlp}(f_2(f_1(\mathbf{X}))) \quad (4)$$

The parameters learned during training of function $f(.)$ will be denoted as $\Theta$. $\Theta$ consists of all learnable parameters of the LSTMs as well as the context vectors in both Level I($f_1$) and Level II($f_2$) Embedding modules, and the parameters of the aforementioned $f_{mlp}$. When training this deep model, parameters are randomly initialized first and input samples are fed through the network. The output of this network is a prediction associated with each sample. The difference between each prediction output $f(\mathbf{X})$ and true label $y$ is fed back into the network through a 'back-propagation' step. The parameters ($\Theta$) are updated in order to minimize a loss function which captures the difference between true labels and predicted values. The loss function $\ell_{Diff}$, on the entire training set of size $N_{samp}$, is defined as:

$$\ell_{Diff} = \frac{1}{N_{samp}} \sum_{n=1}^{N_{samp}} loss(f(\mathbf{X}_{(n)}), y_{(n)}) \quad (5)$$

To train our regression function for the main differential task, we select the squared error loss as the loss function which is defined per-sample as:

$$loss(f(\mathbf{X}^{(n)}), y^{(n)}) = (y^{(n)} - f(\mathbf{X}^{(n)}, \Theta))^2 \quad (6)$$

Thus, we use mean squared error(MSE) $\ell_{Diff}$ for training. The back-propagation step essentially uses stochastic gradient descent (SGD) method to train parameters ([4]). For a set of training samples, instead of calculating true gradient of the objective on all training samples, SGD calculates gradient and updates accordingly on each training sample. Loss $\ell_{Diff}$, calculated previously in Eq. (5) from the training set, is fed into the network and then a gradient descent step is applied to update network parameters $\Theta$ as follows:

$$\Theta \leftarrow \Theta - \eta \frac{\partial L}{\partial \Theta} \quad (7)$$

where $\eta$ is the learning rate parameter and $\frac{\partial L}{\partial \Theta}$ is the gradient. We use the optimizer Adaptive Moment Estimation (ADAM)[19] to train our models. In comparison to SGD in Eq.(7), ADAM computes adaptive learning rates $\eta$, instead of fixed $\eta$ during training, for all parameters from estimates of first and second moments of the gradients. Details of ADAM optimizer are in Appendix Section 1.1.

### 3.6 DeepDiff with Multitasking: Learning Better Representations with Auxiliary Tasks

We then extend the basic DeepDiff mentioned above into a multi-task learning formulation, in order to learn better joint representations informed by auxiliary tasks (details below). Multi-task learning (Multitasking) was initially proposed by Caruana [5] to find common feature representations across multiple relevant tasks. Most of the multitasking studies have focused on neural networks [5], where some hidden layers are shared between various tasks. If different tasks are sufficiently related, multitasking can lead to better generalization. In this paper, we consider two such related tasks as auxiliary tasks for multi-task learning with our DeepDiff main task:

*Cell-Specific Auxiliary – (Auxiliary-Task-A + Auxiliary-Task-B):* We posit the cell-type specific gene expression prediction in each of the two cell-types A and B as the Cell-Specific Auxiliary tasks to the main DeepDiff in the training phase. We call these two tasks in cell-types A and B as Auxiliary-Task-A and Auxiliary-Task-B, respectively. Since our main DeepDiff task uses the log-fold change of the counts of gene expression as the target $y$, we use the log of counts of gene expression in cell-type A and cell-type B as the target value in Auxiliary-Task-A and Auxiliary-Task-B, respectively. To handle zero values of expression, we add 1 to all counts. Besides, we also evaluate using binarized cell-type specific gene expression as the label for the auxiliary tasks (i.e. as classification as opposed to regression. See details in Appendix). For Auxiliary-Task-A, $\mathbf{X}^A$ is passed through two levels of embedding and attention module specific for cell-type A. The output embedding of the Level II Embedding unit $\boldsymbol{v}_A$ is passed through an MLP layer $f_3^A(\boldsymbol{v}_A)$. $f_3^A$ is used to map $\mathbf{X}^A$ to the target value for the gene in Cell-type A. Similarly, for the Auxiliary-Task-B, $\boldsymbol{v}_B$, obtained by passing $\mathbf{X}^B$ through two levels of embedding modules, is passed through an MLP layer $f_3^B(\boldsymbol{v}_B)$ for Cell-type B specific gene expression prediction. By jointly training cell-type specific gene expression with differential gene expression, we can improve the main DeepDiff task performance. We train both of these auxiliary tasks using the sum of the MSE loss averaged over the training set(Eq. (6)) between the gene expression target and predicted values for both cell-types. We denote this cumulative Cell-Specific Auxiliary loss for both cell-types as $\ell_{CellAux}$.

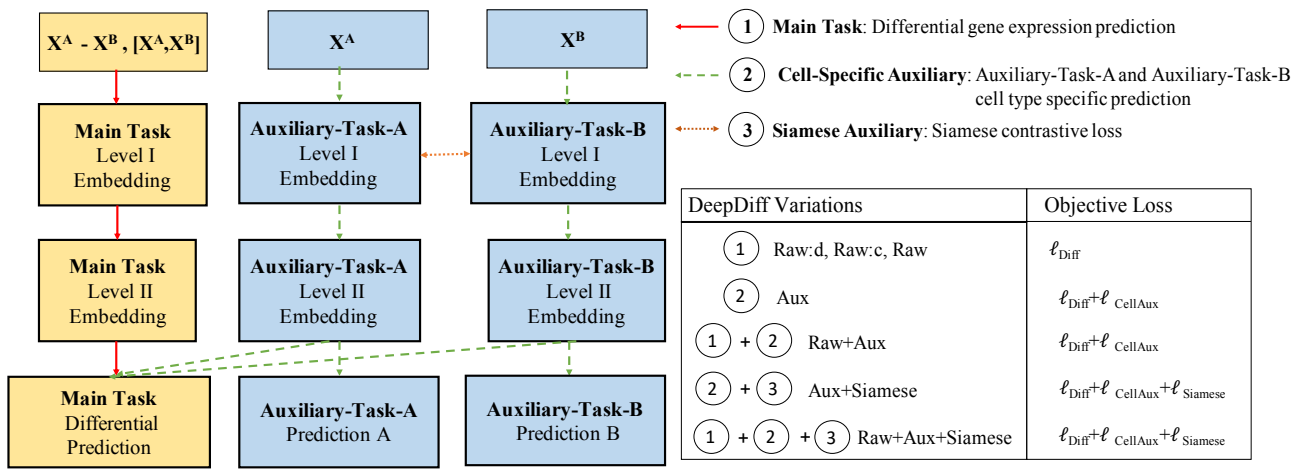

**Fig. 3.** We implement multiple variations of DeepDiff through deep learning based multitasking. Our system includes a set of auxiliary tasks shown as units in this Figure. In details, we use two types of auxiliary tasks coupled with the main DeepDiff task of differential gene expression. The Cell-specific Auxiliary tasks, denoted by Auxiliary-Task-A and Auxiliary-Task-B cell-type specific gene expression prediction. The second Siamese-Auxiliary task uses the siamese contrastive loss at the Level I Embedding outputs. The DeepDiff variations are indicated as combinations of the main task, auxiliary tasks and variations of the input.

*Siamese Auxiliary – with Contrastive Siamese loss:* We use a second auxiliary task to enforce the neighborhood structure of learned representations. This auxiliary task encourages the model to learn embeddings whose neighborhood structures in the model representation space are more consistent with the differential gene expression pattern. This is achieved by a contrastive loss term inspired by the Siamese architecture formulation[15]. In detail, a Siamese architecture consists of two identical networks with shared parameters which accept distinct inputs but are joined by a similarity metric at the output. This similarity metric, used in the output loss of the network, coupled with shared weights encourages 'similar' inputs to map to nearby points in the output representation space and 'dissimilar' inputs to map to distant points in the representation space. We extend this notion of similarity and dissimilarity to the differential gene expression case. We consider the histone modification profiles $\boldsymbol{X}^A$ and $\boldsymbol{X}^B$ of two differentially expressed genes(upregulated or downregulated) to be 'different' and 'similar' for genes not differentially expressed. We denote 'dissimilar' with label $S = 1$ and 'similar' with label $S = 0$. If log change in differential gene expression $<= -2$(downregulated) or $>= 2$(upregulated), we label it as differentially regulated($S = 1$). Otherwise, we label the training sample with $S = 0$. For this auxiliary task, we use Level I Embedding modules as Siamese twin networks in DeepDiff variations. For example, $\mathbf{X}^A$ is passed through two levels of embedding and attention modules specific to cell-type A. Similarly, $\mathbf{X}^B$ is passed through two levels of embedding and attention modules. For the Siamese contrastive loss, we use the Level I embeddings for Cell-type A and B: $f_1^A$ and $f_1^B$. We use the following siamese contrastive loss[15], denoted as $\ell_{Siamese}$, at the output of the Level I Embedding units $f_1^A$ and $f_1^B$ to train this auxiliary task:

$$\ell_{Siamese} = (1-S) \times \frac{1}{2} \times R + S \times \frac{1}{2} \max(0, m-R)^2 \quad (8)$$

Here, R represents:

$$R = \sqrt{(f_1^A(\boldsymbol{X}^A) - f_1^B(\boldsymbol{X}^B))^2} \quad (9)$$

Many possible variations of DeepDiff exist through the various combinations of the auxiliary tasks in the multi-tasking framework as well as the raw HM features. For example, we can use all three auxiliary tasks to multi-task with the main task. This means the sum of the main and auxiliary task losses is used as part of the training objective: (1) Differential expression prediction loss for the main task ($\ell_{Diff}$), (2) Auxiliary task loss ($\ell_{CellAux}$) from the Cell-specific prediction tasks, and (3) Contrastive Siamese Loss ($\ell_{Siamese}$) from the Siamese-Auxiliary Task. The network is trained using similar steps outlined in the main DeepDiff task with the sum of these losses as training objective.

In our experiments, we have evaluated the following DeepDiff variations:

- **(Raw:d)** Raw Difference Features ;
- **(Raw:c)** Concatenation of Raw HM features ;
- **(Raw)** Concatenation and difference of raw HM features;
- **(Aux)** Auxiliary Embeddings as Features;
- **(Raw+Aux)** Concatenation and Difference of HMs + Embeddings from Auxiliary tasks;
- **(Aux+Siamese)** Auxiliary Features with Siamese Contrastive Loss;

- **(Raw+Aux+Siamese)** Raw and Auxiliary Features with Siamese Contrastive Loss

Figure 3 and Appendix Table 1 show the different auxiliary tasks and resulting different DeepDiff variations through various combinations with the raw HM features. Due to space limitation, we have the detailed description of each variation in Appendix Section 1.3. In the rest of the paper, we use the short names enclosed in parantheses above to refer to each variation.

## 4 Experimental Setup and Results

### 4.1 Dataset

We downloaded gene expression and HM signal data of five core histone modification signals for ten different cell types from the REMC database([22]). Appendix Table 5 summarizes the IDs and information of the ten cell-types we use that have been extensively profiled by the Roadmap Epigenomics Project. Appendix Table 4 describes the five core histone modifications marks we use along with their known important roles in gene regulation. For the regression labels, we use the log fold change of raw counts in the two cell-types under consideration as the target label for the main task and the log of the raw counts in each cell type for the two cell-specific auxiliary tasks. In total, we apply DeepDiff variations and baselines(see Section 4.2) on ten pairs of cell-types from REMC. Appendix Table 6 provides a list of the ten cell-type pairs. For each cell type pair, we have a sample set of total 18460 genes. This set was divided into 3 separate folds: training (10000 genes), validation (2360 genes) and test (6100 genes) folds.

### 4.2 Baselines

We compare DeepDiff to two variations of the Support Vector Regression (SVR)[6], the ReliefF based feature selection followed by Random Forest Regression[23] and the AttentiveChrome([29]). In details:

- *Single Layer SVR ([6]):* The authors selected 160 bins from regions flanking each gene TSS and TTS. Each bin uses a separate SVR, resulting in 160 different bin-specific SVR models. The radial basis kernel is used for the SVR. [6] proposed to use the best-bin strategy. Therefore, by using cross-validation, we pick the best bin based on Pearson Correlation Coefficient(PCC) used by [6]. The best bin model is then used for prediction on the test set.
- *Two Layer SVR ([6]):* The two-layer model in [6] seeked to combine the signals of all HMs across all the 160 bins. In the first layer, it predicts expression levels in each of the bins using a bin-specific SVR model in each individual bin. Then the expression levels predicted by each bin are combined in the second layer using another SVR model to make a final prediction. The radial basis kernel is used for both layers of the SVR.
- *ReliefF Feature Selection + Random Forest[23]:* We implement the best performing combination in [23]: ReliefF algorithm for feature selection followed by random forest. While [23] treat the problem as a binary classification with two classes of upregulated vs downregulated genes, we used this baseline for our regression formulation. The number of features for ReliefF is selected from the set of $\{50, 70, 100\}$. The number of trees for Random Forest is selected from the set of $\{10, 50, 100, 150, 200\}$.
- AttentiveChrome([29]): We also compare our models to the differential patterns derived from predictions made by AttentiveChrome. We train AttentiveChrome for cell type specific gene expression as a regression task. Then we calculate the differential gene expression prediction as the log fold change between the predicted expression from the two trained models that are specific to the two cell-types under consideration. In detail, for each pair of cell-types, we train two cell-specific AttentiveChrome models independently from each other. We then use the cell type specific predictions of each model to calculate differential gene expression. Clearly, this baseline is not an end-to-end solution for differential gene expression prediction.

We implemented SVR and Random Forest baselines using the scikit-learn ([27]) package. We implemented AttentiveChrome and DeepDiff models in Pytorch. We use Pearson Correlation Coefficient to evaluate our models. We train all the variations of DeepDiff (summarized in Appendix Table 1) were trained using our training set and tune the hyperparameter on the validation set. The best performing models were then evaluated on the test set. The details about evaluation metric and hyperparameters for DeepDiff and baselines are in the Appendix Section 1.5.

### 4.3 Performance Evaluation

Figure 4 shows the Pearson Correlation Coefficient (y-axis) for all DeepDiff variations versus the baselines. The x-axis shows the ten cases (cell-type pairs) in our experiments. The deep learning based models outperform both the SVR as well as Random Forest baselines. The two-layer SVR model performs better than one layer SVR. When comparing DeepDiff variations to the AttentiveChrome baseline, DeepDiff also outperforms, indicating the need for modeling differential gene expression prediction. Among the DeepDiff variations, the *Raw:d* model performs the worst in 9 out of the 10 cases. This indicates that simply taking the difference of the HM signals is not enough to model the combinatorial interactions of HMs for differential regulation. Instead, using all HM features, instead of only the difference, is clearly helpful, as indicated by the higher PCC of *Raw:c* in comparison to *Raw:d* across all 10 cases. The results of *Raw+Aux* show that adding cell-type specific gene expression prediction in the two cell-types as auxiliary tasks clearly helps in improving the prediction performance. Combining *Aux* and *Raw* features gives better performance than only *Aux* and *Raw* in 7 and 6 out of the 10 cell type pairs, respectively. *Aux+Siamese* is the best performing model in 5 out of the 10 cases. This shows that adding the Siamese-based contrastive loss improves the prediction performance. Table 1 shows the mean and median of the relative performance(%) with respect to PCC when being compared to the best performing baselines: two-layer SVR and AttentiveChrome. For example, as shown in Table 1, when averaged across the 10 cases, combining the raw and auxiliary features (*Raw+Aux* variation)results in a relative PCC with respect to the two-layer SVR baseline of 162.23%.

### 4.4 Interpreting differential regulation using attention

Finally, we analyze the attention weights of Level II Embedding for one of the best performing case: cell type pair E116 and E123. Here, E116 represents 'normal' blood cell (GM12878) whereas E123 represents the leukemia cell (K562). We aim to validate that the learned attention weights can provide some insights into the differential gene regulation across these two selected cell types: a normal and a diseased (cancer) cell state. We only use the attention weights obtained on the test set for this analysis. First, we get a list of down-regulated genes (with log fold change $< -8.0$) and a list of up-regulated genes (with log fold change $> 8.0$) from the test set. Figure 5 plots the average attention weights of each of the 5 HMs when considering all of our selected up-regulated(black bars) and down-regulated genes (gray bars). We observe that for the top upregulated genes, H3K4me1 (enhancer associated) and H3K4me3 (promoter associated) get the highest weights (i.e. contributing more importance). This is consistent with observations by [14] that upregulated genes are more enriched for H3K4me1 and H3K4me3 when under cancer condition.

For top down-regulated genes (Figure 5), H3K27me3 (Polycomb Repression associated) gets a comparatively higher weight (the second highest attention weight). [14] also reported this trend showing that down-regulated genes are more enriched with the repressive H3K27me3 under the cancer condition whereas up-regulated genes do not show variation of this HM between normal and cancer conditions. Figure 5 shows that H3K9me3 (heterochromatin linked) gets low attention weights for both up-regulated and down-regulated genes. [14] also reported this trend that the divergence of expression is less likely due to this HM. We want to emphasize again that our model learns the importance of HMs for the differentially expressed genes in an end-to-end manner.

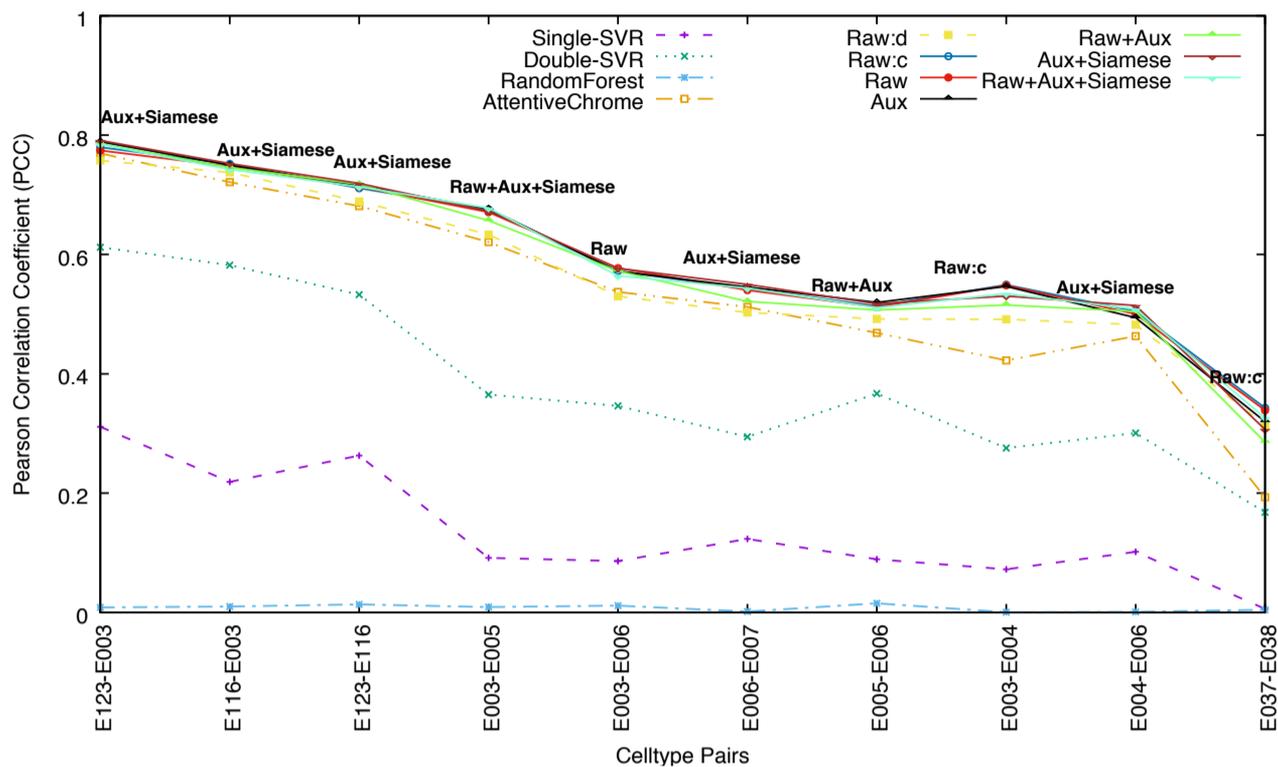

**Fig. 4.** Pearson Correlation Coefficient (PCC) for all DeepDiff variations along with the baselines for each cell-type pair(x-axis). The best performing DeepDiff model variation is indicated by the text label for each case (cell-type pair).

| Method | two-layer SVR | | AttentiveChrome | |
|---|---|---|---|---|
| | Mean | Median | Mean | Median |
| *Raw:d* | 153.72 | 156.81 | 108.90 | 102.14 |
| *Raw:c* | 163.59 | 166.64 | 115.75 | 107.64 |
| *Raw* | 162.94 | 166.55 | 115.26 | 107.71 |
| *Aux* | 157.64 | 166.58 | 111.20 | 106.13 |
| *Raw+Aux* | 162.23 | 164.73 | 114.61 | 106.62 |
| *Aux+Siamese* | 161.85 | 168.90 | 114.19 | 107.87 |
| *Raw+Aux+Siamese* | 161.95 | 166.07 | 114.44 | 107.56 |

Table 1. Mean and Median of the relative performance (%) with respect to Pearson Correlation Coefficient(PCC) when comparing DeepDiff models to the best-performing baselines: two-layer SVR and AttentiveChrome across all ten cell-type pairs.

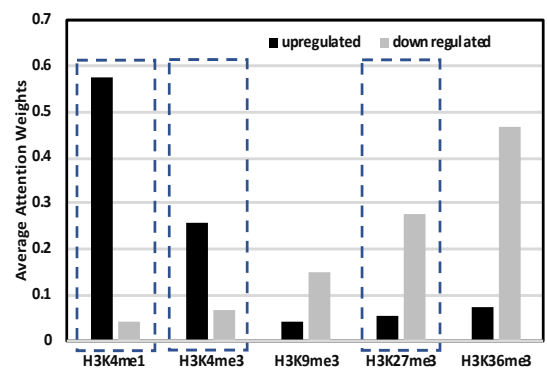

**Fig. 5.** HM attention weights at Level II Embedding for one of the best performing cell type pair - E116 and E123. Here, E116 represents 'normal' blood cell (GM12878) whereas E123 represents 'cancer' or leukemia cell (K562). We plot the average attention weights for all 5 HMs across down-regulated (log fold change $< -8.0$) and up-regulated genes (log fold change $> 8.0$). H3K4me1 and H3K4me3 get the highest weights in upregulated genes and comparatively lower weights in downregulated genes, indicating the importance of these histone modifications for upregulation. Similarly, H3K27me3 gets a comparatively higher weight (second highest weight) in downregulated genes than the same histone modification in upregulated genes. Both these observations have been confirmed previously by [14] through HM enrichment analysis across the two cell types.

## 5 Conclusion

We have presented DeepDiff, a deep learning framework for differential gene expression prediction using histone modifications (HMs). DeepDiff is an attention-based deep learning architecture designed to understand how different HMs work together to influence changes in expression patterns of a gene between two different cell types. DeepDiff uses a modular architecture to represent the spatially structured and long-range HM signals. It incorporates a two-level attention mechanism that gives it the ability to find salient features at the bin level as well as the HM level. Additionally, to deal with fewer differentially expressed genes between two cell types, we design a novel multi-task framework to use the cell-type-specific prediction network as auxiliary tasks to regularize our primary task of differential expression prediction. We also incorporate a Siamese contrastive loss term to further improve the learned representations. For the future work, we will evaluate the performance and the attention scores of DeepDiff on more cell type pairs. We will incorporate additional epigenomic signals that may relate to differential gene expression. We would also like to explore different ways to interpret and validate the attention weights. In summary, leveraging deep learning's ability to extract rich representations from data can enhance our understanding of gene regulation by HMs, thus enabling insights into principles of gene regulation through epigenetic factors.

## References

[1] Jimmy Ba, Volodymyr Mnih, and Koray Kavukcuoglu. Multiple object recognition with visual attention.

[2] Dzmitry Bahdanau, Kyunghyun Cho, and Yoshua Bengio. Neural machine translation by jointly learning to align and translate. *arXiv preprint arXiv:1409.0473*, 2014.

[3] Andrew J Bannister and Tony Kouzarides. Regulation of chromatin by histone modifications. *Cell research*, 21(3):381–395, 2011.

[4] Léon Bottou. Stochastic learning. In *Advanced lectures on machine learning*, pp. 146–168. Springer, 2004.

[5] Rich Caruana. Multitask learning. *Machine learning*, 28(1):41–75, 1997.

[6] Chao Cheng and Mark Gerstein. Modeling the relative relationship of transcription factor binding and histone modifications to gene expression levels in mouse embryonic stem cells. *Nucleic acids research*, 40(2):553–568, 2011.

[7] Jan K Chorowski, Dzmitry Bahdanau, Dmitriy Serdyuk, Kyunghyun Cho, and Yoshua Bengio. Attention-based models for speech recognition. In C. Cortes, N. D. Lawrence, D. D. Lee, M. Sugiyama, and R. Garnett, editors, *Advances in Neural Information Processing Systems 28*, pp. 577–585. Curran Associates, Inc., 2015.

[8] Maurizio Corbetta and Gordon L Shulman. Control of goal-directed and stimulus-driven attention in the brain. *Nature reviews neuroscience*, 3(3):201–215, 2002.

[9] Ivan G Costa, Helge G Roider, Thais G Rego, and Francisco AT Carvalho. Predicting gene expression in t cell differentiation from histone modifications and transcription factor binding affinities by linear mixture models. *BMC bioinformatics*, 12(1):1, 2011.

[10] Xianjun Dong, Melissa C Greven, Anshul Kundaje, Sarah Djebali, James B Brown, Chao Cheng, Thomas R Gingeras, Mark Gerstein, Roderic Guigó, Ewan Birney, et al. Modeling gene expression using chromatin features in various cellular contexts. *Genome Biol*, 13(9):R53, 2012.

[11] Gerda Egger, Gangning Liang, Ana Aparicio, and Peter A Jones. Epigenetics in human disease and prospects for epigenetic therapy. *Nature*, 429(6990):457, 2004.

[12] Marco Frasca and Giulio Pavesi. A neural network based algorithm for gene expression prediction from chromatin structure. In *Neural Networks (IJCNN), The 2013 International Joint Conference on*, pp. 1–8. IEEE, 2013.

[13] Elizabeta Gjoneska, Andreas R Pfenning, Hansruedi Mathys, Gerald Quon, Anshul Kundaje, Li-Huei Tsai, and Manolis Kellis. Conserved epigenomic signals in mice and humans reveal immune basis of alzheimer's disease. *Nature*, 518(7539):365, 2015.

[14] Laura Grégoire, Annabelle Haudry, and Emmanuelle Lerat. The transposable element environment of human genes is associated with histone and expression changes in cancer. *BMC genomics*, 17(1):588, 2016.

[15] Raia Hadsell, Sumit Chopra, and Yann LeCun. Dimensionality reduction by learning an invariant mapping. In *Computer vision and pattern recognition, 2006 IEEE computer society conference on*, volume 2, pp. 1735–1742. IEEE, 2006.

[16] Bich Hai Ho, Rania Mohammed Kotb Hassen, and Ngoc Tu Le. Combinatorial roles of dna methylation and histone modifications on gene expression. In *Some Current Advanced Researches on Information and Computer Science in Vietnam*, pp. 123–135. Springer, 2015.

[17] Sepp Hochreiter and Jürgen Schmidhuber. Long short-term memory. volume 9, pp. 1735–1780. MIT Press, 1997.

[18] Rosa Karlić, Ho-Ryun Chung, Julia Lasserre, Kristian Vlahoviček, and Martin Vingron. Histone modification levels are predictive for gene expression. *Proceedings of the National Academy of Sciences*, 107(7):2926–2931, 2010.

[19] Diederik Kingma and Jimmy Ba. Adam: A method for stochastic optimization. 2014.

[20] Christoph M Koch, Robert M Andrews, Paul Flicek, Shane C Dillon, Ulaş Karaöz, Gayle K Clelland, Sarah Wilcox, David M Beare, Joanna C Fowler, Phillippe Couttet, et al. The landscape of histone modifications across 1% of the human genome in five human cell lines. *Genome research*, 17(6):691–707, 2007.

[21] Igor Kononenko, Edvard Šimec, and Marko Robnik-Šikonja. Overcoming the myopia of inductive learning algorithms with relieff. *Applied Intelligence*, 7(1):39–55, 1997.

[22] Anshul Kundaje, Wouter Meuleman, Jason Ernst, Misha Bilenky, Angela Yen, Alireza Heravi-Moussavi, Pouya Kheradpour, Zhizhuo Zhang, Jianrong Wang, Michael J Ziller, et al. Integrative analysis of 111 reference human epigenomes. *Nature*, 518(7539):317–330, 2015.

[23] Jeffery Li, Travers Ching, Sijia Huang, and Lana X Garmire. Using epigenomics data to predict gene expression in lung cancer. In *BMC bioinformatics*, volume 16, p. S10. BioMed Central, 2015.

[24] Minh-Thang Luong, Hieu Pham, and Christopher D. Manning. Effective approaches to attention-based neural machine translation. In *Empirical Methods in Natural Language Processing (EMNLP)*, pp. 1412–1421, Lisbon, Portugal, September 2015. Association for Computational Linguistics.

[25] Michele Meisner and David M Reif. Computational methods used in systems biology. In *Systems Biology in Toxicology and Environmental Health*, pp. 85–115. Elsevier, 2015.

[26] Volodymyr Mnih, Nicolas Heess, Alex Graves, and others. Recurrent models of visual attention. In *Advances in neural information processing systems*, pp. 2204–2212.

[27] F. Pedregosa, G. Varoquaux, A. Gramfort, V. Michel, B. Thirion, O. Grisel, M. Blondel, P. Prettenhofer, R. Weiss, V. Dubourg, J. Vanderplas, A. Passos, D. Cournapeau, M. Brucher, M. Perrot, and E. Duchesnay. Scikit-learn: Machine learning in Python. *Journal of Machine Learning Research*, 12:2825–2830, 2011.

[28] Ritambhara Singh, Jack Lanchantin, Gabriel Robins, and Yanjun Qi. Deepchrome: deep-learning for predicting gene expression from histone modifications. *Bioinformatics*, 32(17):i639–i648, 2016.

[29] Ritambhara Singh, Jack Lanchantin, Arshdeep Sekhon, and Yanjun Qi. Attend and predict: Understanding gene regulation by selective attention on chromatin. In I. Guyon, U. V. Luxburg, S. Bengio, H. Wallach, R. Fergus, S. Vishwanathan, and R. Garnett, editors, *Advances in Neural Information Processing Systems 30*, pp. 6785–6795. Curran Associates, Inc., 2017.

[30] Ilya Sutskever, Oriol Vinyals, and Quoc V Le. Sequence to sequence learning with neural networks. In *Advances in neural information processing systems*, pp. 3104–3112, 2014.

[31] Oriol Vinyals, Meire Fortunato, and Navdeep Jaitly. Pointer networks. In C. Cortes, N. D. Lawrence, D. D. Lee, M. Sugiyama, and R. Garnett, editors, *Advances in Neural Information Processing Systems 28*, pp. 2692–2700. Curran Associates, Inc., 2015.

[32] Nan-ping Weng, Yasuto Araki, and Kalpana Subedi. The molecular basis of the memory t cell response: differential gene expression and its epigenetic regulation. *Nature Reviews Immunology*, 12(4):306, 2012.

[33] Huijuan Xu and Kate Saenko. Ask, attend and answer: Exploring question-guided spatial attention for visual question answering. In *ECCV*, 2016.

[34] Kelvin Xu, Jimmy Ba, Ryan Kiros, Kyunghyun Cho, Aaron C Courville, Ruslan Salakhutdinov, Richard S Zemel, and Yoshua Bengio. Show, attend and tell: Neural image caption generation with visual attention. In *ICML*, volume 14, pp. 77–81, 2015.

[35] Zichao Yang, Diyi Yang, Chris Dyer, Xiaodong He, Alex Smola, and Eduard Hovy. Hierarchical attention networks for document classification. 2016.

[36] Li Yao, Atousa Torabi, Kyunghyun Cho, Nicolas Ballas, Christopher Pal, Hugo Larochelle, and Aaron Courville. Describing videos by exploiting temporal structure. In *Computer Vision (ICCV), 2015 IEEE International Conference on*. IEEE, 2015.

# 1 Appendix

## 1.1 Background: ADAM optimizer

ADAM[7] computes adaptive learning rates $\eta$ for all parameters from estimates of first and second moments of the gradients. The first moment($\hat{m}_t$) involves the exponentially decaying average of the previous gradients and the second moment($\hat{v}_t$) involves exponentially decaying average of the previous squared gradients. The update rule for epoch $t$ during training is:

$$\Theta_t \leftarrow \Theta_{t-1} - \frac{\eta}{\sqrt{\hat{v}_t} + \epsilon}\hat{m}_t \quad (1)$$

Here, $\epsilon$ is a very small number to prevent division by zero.

## 1.2 Background: Siamese Network in Deep Learning

The Siamese architecture has been used in many real applications, like face recognition [8] and dimension reduction [4]. A Siamese network contains two copies of a deep neural network(DNN) sharing the same weights($W$). Figure 1 shows a general schema of a siamese architecture. Inputs are pairs of samples $X^A$ and $X^B$. The two twin networks are tied by a distance measure($D_W(X^A, X^B)$) computed at the output representations of the two twin networks. A meaningful mapping maps similar input vectors to nearby points on the output manifold and dissimilar vectors to distant points. Inputs are pairs of samples. By forwarding a pair of similar samples into the Siamese network and penalizing the outputs (distance) of the pair, we can intuitively limit the distance between two similar samples in the learned embedding space to be small.

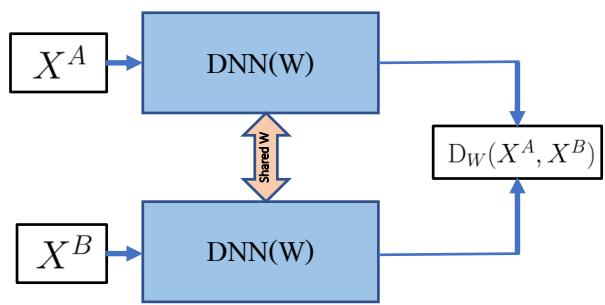

**Fig. 1.** Schematic of a general Siamese Network. Inputs are pairs of samples. By forwarding a pair into the Siamese network and penalizing the outputs of the pair, this training intuitively limits the $D_W$ distance between two similar samples to be small. Backpropagation is used to train the network.

## 1.3 DeepDiff Variations Tried in our Experiments

We focus on the predictive modeling of *differential* gene expression given the histone modification profiles of a gene in two cell-types. To improve the prediction of differential gene expression, we use two types of auxiliary information. We use the cell-type specific expression prediction as an auxiliary task to the main task of differential gene expression prediction. Additionally, we also introduce a contrastive loss term as an auxiliary regularization term to further aid differential gene expression prediction. Combining these different auxiliary terms helps the model build powerful representations to improve differential gene expression prediction performance. Figure 3 in Section 3.6 presents an overview of our strategy. We use a number of variations of the DeepDiff model:

*Raw Difference Features (Raw:d)* First, we predict differential gene expression using the difference of the corresponding HM signals $X = X^A - X^B$. We use $X$ directly as input to the previously described Level I Embedding module $f_1$. The outputs from the Level I Embedding module are used as input to the Level II Embedding module $f_2$. This embedding $v$ is passed through a linear layer for prediction. Intuitively, this is exactly like the AttentiveChrome model with input $X = X^A - X^B$(shown in Figure 2 in Section 3.4).

*Concatenation of Raw HM features (Raw:c)* In this model, we treat the HM level signals a gene from the two cell-types as different HM features. We concatenate the HM profiles from the two cell-types into a single matrix $X = [X^A, X^B]$ of size $(2 \times M) \times T$. This is used as input to the Level I Embedding module $f_1$ followed by the Level II Embedding module $f_2$. We use a Level I Embedding module that has one LSTM for each HM (from both cell-types), bin level attention weights $\alpha_{jt}$, $j \in [1 \ldots 2 \times M]$ and $t \in [1 \ldots T]$ followed by a Level II Embedding module with HM level attention weights $\beta_j$ $j \in [1 \ldots 2 \times M]$. Similar to *Raw:d* model, we only predict differential expression.

*Concatenation and difference of raw HM features(Raw):* In addition to the concatenated HM features, this variation uses an additional set of features corresponding to the difference of the HM profiles: $X^A - X^B$. Thus, the input matrix is now $X = [X^A, X^B, X^A - X^B]$, a $(3 \times M) \times T$ matrix. We use this matrix as the input to the Level I Embedding module $f_1$ followed by the Level II Embedding module $f_2$. This Level II Embedding $v$ is passed through a linear layer for prediction.

*Adding Features from Auxiliary Tasks and Auxiliary Contrastive Loss:* We propose using individual gene expression prediction as an auxiliary task to help the harder task of differential gene expression prediction. For this purpose, we propose the following variations:

*Concatenation and Difference of HMs + Auxiliary features(Raw+Aux):* This model aims at better feature representations for the *Raw* model. For this purpose, we use $X = [X^A, X^B, X^A - X^B]$ as the input to a Level I Embedding module $f_1^d$, followed by a Level II Embedding module $f_2^d$ for the DeepDiff main task. We add cell-type specific gene expression prediction for cell-type A and B as the Cell-Specific Auxiliary task(Auxiliary-Task-A and Auxiliary-Task-B, respectively). For this purpose, another Level I Embedding module $f_1^A$ takes as input matrix $X^A$ corresponding to the HM profile for cell-type $A$. This is followed by a Level II Embedding module $f_2^A$ for cell-type A. Similarly, we use another Level I Embedding module $f_1^B$ followed by the Level II Embedding module $f_2^B$ for cell-type B. To leverage the information from the cell-type specific expression prediction tasks, additional auxiliary features are provided to the Level II Embedding module $f_2^d$. Concretely, in addition to the outputs of $f_1^d$, the Level II Embedding module $f_2^d$ also takes as features outputs from the Level I Embedding modules $f_1^A$ and $f_1^B$. Thus, $f_2^d$ receives as input the output representations from both the $f_1^A$ and $f_1^B$ Level I Embedding units concatenated after the $f_1^d$ Level I Embedding module outputs. Both the Cell-Specific Auxiliary task and the main difference tasks are trained end to end together.

*Only Auxiliary Embedding as Features(Aux):* For this variation, at the first level we use two Level I Embedding modules $f_1^A$ and $f_1^B$ corresponding to each cell-type. This is followed by two Level II Embedding modules $f_2^A$ and $f_2^B$ that take as input $f_1^A(X^A)$ and $f_1^B(X^B)$ respectively. The output of the Level II Embedding modules gives two final auxiliary embeddings $v_A$, and $v_B$ (Auxiliary-Task-A Embedding and Auxiliary-Task-B Embedding, respectively). These auxiliary embeddings are concatenated, $v = [v_A, v_B]$, and used as input to an MLP for the final prediction. For the auxiliary task predictions, the output $v_A$ is passed through a linear layer for the prediction for cell-type $A$. Similarly, $\boldsymbol{v}_B$ is used as input to a linear layer for the prediction for cell-type $B$.

| Model | Input Features | Auxiliary Task | Target Labels | Level I Embedding | Level II Embedding | Loss |
|---|---|---|---|---|---|---|
| *Raw:d* | $\boldsymbol{X}^A - \boldsymbol{X}^B$ | - | differential expression | $f_1$ | $f_2$ | $\ell_{Diff}$ |
| *Raw:c* | $[\boldsymbol{X}^A, \boldsymbol{X}^B]$ | - | differential expression | $f_1$ | $f_2$ | $\ell_{Diff}$ |
| *Raw* | $[\boldsymbol{X}^A, \boldsymbol{X}^B, \boldsymbol{X}^A - \boldsymbol{X}^B]$ | - | differential expression | $f_1$ | $f_2$ | $\ell_{Diff}$ |
| *Raw+Aux* | $[\boldsymbol{X}^A, \boldsymbol{X}^B, \boldsymbol{X}^A - \boldsymbol{X}^B]$ and $\boldsymbol{X}^A, \boldsymbol{X}^B$ | Cell-Specific Auxiliary | differential expression, gene expression A, gene expression B | $f_1^d, f_1^A, f_1^B$ | $f_2^d, f_2^A, f_2^B$ | $\ell_{Diff} + \ell_{CellAux}$ |
| *Aux* | $\boldsymbol{X}^A, \boldsymbol{X}^B$ | Cell-Specific Auxiliary | differential expression, gene expression A, gene expression B | $f_1^A, f_1^B$ | $f_2^A, f_2^B$ | $\ell_{Diff} + \ell_{CellAux}$ |
| *Aux+Siamese* | $\boldsymbol{X}^A, \boldsymbol{X}^B$ | Cell-Specific Auxiliary + Siamese Auxiliary | differential expression, gene expression A, gene expression B | $f_1^A, f_1^B$ (shared weights) | $f_2^A, f_2^B$ | $\ell_{Diff} + \ell_{CellAux} + \ell_{Siamese}$ |
| *Raw+Aux+Siamese* | $[\boldsymbol{X}^A, \boldsymbol{X}^B, \boldsymbol{X}^A - \boldsymbol{X}^B], \boldsymbol{X}^A, \boldsymbol{X}^B$ | Cell-Specific Auxiliary + Siamese Auxiliary | differential expression, gene expression A, gene expression B | $f_1^d, f_1^A, f_1^B$ (shared weights for A and B) | $f_2^d, f_2^B$ | $\ell_{Diff} + \ell_{CellAux} + \ell_{Siamese}$ |

Table 1. DeepDiff Variations in detail: The columns represent (a) different combinations of input features, (b) the auxiliary tasks used in the multitasking framework (Cell-Specific Auxiliary includes both the Auxiliary-Task-A and Auxiliary-Task-B), (c) the corresponding target labels for the tasks, and (d),(e) the model architecture of the variations: $f_1$ represents Level I Embedding module, and $f_2$ represents Level II Embedding module, and (f) the corresponding loss used to train the models.

*Siamese Auxiliary with Siamese Contrastive Loss(Aux+Siamese):* Using the siamese contrastive loss formulation[4], we introduce a notion of similarity and dissimilarity based on a gene's differential gene expression. We consider the histone modification profiles $\boldsymbol{X}^A$ and $\boldsymbol{X}^B$ of two differentially expressed genes(upregulated or downregulated) to be 'different'($S = 1$) and 'similar'($S = 0$) for genes not differentially expressed. We introduce a contrastive loss term $\ell_{Siamese}$ as a regularizer, based on whether a gene is differentially regulated or not at the output embedding of the Level I Embedding unit $f_1$. We use the following formulation $\ell_{Siamese}$:

$$\ell_{Siamese} = (1 - S) \times \frac{1}{2} \times R + S \times \frac{1}{2}\max(0, m - R)^2 \quad (2)$$

where:

$$R = \sqrt{(f_1^A(\boldsymbol{X}^A) - f_1^B(\boldsymbol{X}^B))^2} \quad (3)$$

In Eq. 2, $m > 0$ is the margin in the contrastive loss and $S$ indicates similarity or dissimilarity of the inputs i.e., $S = 1$ if the gene is differentially expressed, and $S = 0$ if not differentially regulated. Contrastive Loss encourages 'similar' inputs to map to nearby points in the output representation space and 'dissimilar' inputs to map to distant points in the representation space. We use this $\ell_{Siamese}$ as a regularizer. We classify genes based on log change in differential gene expression$<= -2$(downregulated) or differential gene expression$>= 2$(upregulated) as differentially regulated($S = 1$) and log change in $-2 <=$differential gene expression$<= 2$ as $S = 0$. For this model, we use the Level I embedding unit as Siamese twin networks, i.e. $f_1^A$ and $f_1^B$ share their weights, while $f_2^A$ and $f_2^B$ (similar to the *aux* model) do not share weights.

*Raw and Auxiliary Features with Siamese Contrastive Loss(Raw+Aux+Siamese):* We further add the above contrastive loss formulation to the *Raw+Aux* model. We use the Level I Embeddings $f_1^A$ and $f_1^B$ as Siamese twin networks that share weights, and use the concatenation of the output Level I embeddings for the contrastive loss $\ell_{Siamese}$ in Eq. 2. In addition to $f_1^A$ and $f_1^B$, we use $f_1^d$ for the *Raw* features, similar to *Raw+Aux* model.

For the models with auxiliary tasks, *Raw+Aux* and *Aux*, we use the total loss $\ell = \ell_{Diff} + \ell_{CellAux}$. For *Aux+Siamese*, we use $\ell = \ell_{Diff} + \ell_{CellAux} + \ell_{Siamese}$. For the *Raw*, *Raw:c* and *Raw:d* models, we only use $\ell_{Diff}$ for optimizing the network. Table 1 shows the DeepDiff variations with corresponding architecture, target labels and loss variations. Figure 3 in Section 3.6 presents the variations as a combination of the DeepDiff main and Cell-Specific Auxiliary tasks.

### 1.4 Related Work

Table 2 compares DeepDiff with all the related studies discussed in Section 2 for the task of quantifying gene expression using HMs.

### 1.5 More about experimental setup

*DeepDiff and baseline hyperparameters:* For Level I Embedding, we use bidirectional LSTMs with hidden state size $D = 32$. Similarly, for bidirectional LSTMs in Level II Embedding modules, we use the hidden state size of 16. Since we implement a bi-directional LSTM, this results in each hidden state at Level I Embedding hidden state $\boldsymbol{h}_{jt}$ of size 64 and Level II Embedding hidden state $s_j$ of size 32. Accordingly, we set the context vectors, $\boldsymbol{Wb}_j$ and $\boldsymbol{W}_h$, to size 64 and 32, respectively. We also use dropout, a regularization technique based on randomly dropping units from DNNs during training to prevent overfitting. We use a dropout probability of 0.5 for our experiments. We use hyperparameter $m = 2.0$ in our experiments for the *Aux+Siamese* and *Raw+Aux+Siamese* models with Siamese Auxiliary task (Equation (2)). For both the single and two-layer SVR models, we used cross-validation on varying hyperparameter values of $C \in \{0.1, 1, 10, 100\}$. We used radial basis kernel for SVR models. For the rest of the parameters, we used default settings in sklearn.

*Evaluation Metric:* We use Pearson Correlation Coefficient (PCC) to evaluate all our variations and baselines. PCC is a measure of the linear correlation between two continuous variables (predicted and target values in our experiments). It ranges between 1 and $-1$, where 1 is total positive linear correlation, 0 is no linear correlation, and $-1$ is total negative linear correlation.

### 1.6 More possible experiments: Classification as Cell-Specific Auxiliary Task

We also evaluate using classification labels for the Cell-Specific Auxiliary Task as opposed to regression. To formulate the labels in cell-type specific gene expression prediction in each cell-type as binary classification, we follow AttentiveChrome. In detail, for each cell type, we choose the cell type specific median of the rpkm gene expression as the threshold to classify the expression as 1 or $-1$. We use the log fold change in rpkm gene expression values as the regression label for the differential expression task. If the auxiliary task is classification, $v'_A$, defined in Appendix Section 1.3 for Cell-Specific Auxiliary task, will be fed to a softmax output layer. To train this classification auxiliary task, we minimize the negative log likelihood loss. Figure 2 shows the PCC for all model variations for $classification$ of cell-type specific gene expression as auxiliary tasks and Table 3 shows the relative performance (%) with respect to Pearson Correlation Coefficient(PCC) when comparing *Aux* and *Raw+Aux* models to two-layer SVR. Because rpkm is a cell type specific normalization, we use rpkm as the target label in this case for consistency with the labels for Cell-Specific Auxiliary tasks.

| Computational Study | Differential | Unified | Non-linear | Bin-Info | Representation Learning | | Feature Inter. | Interpretable | Output |
|---|---|---|---|---|---|---|---|---|---|
| | | | | | Neighbor Bins | Whole Region | | | |
| Linear Regression ([6]) | × | × | × | × | × | ✓ | × | ✓ | Regression |
| SVR (single layer) ([1]) | ✓ | × | ✓ | Bin-specific | × | ✓ | ✓ | × | Regression |
| SVR (two layer) ([1]) | ✓ | × | ✓ | × | × | ✓ | ✓ | × | Regression |
| SVM ([2]) | ✓ | × | ✓ | Bin-specific | × | ✓ | ✓ | × | Classification |
| Random Forest ([3]) | × | × | ✓ | Best-bin | × | ✓ | × | × | Classification/Regression |
| ReliefF+Random Forest ([10]) | ✓ | × | ✓ | × | × | ✓ | × | × | Classification |
| Rule Learning ([5]) | × | × | ✓ | × | × | ✓ | ✓ | ✓ | No prediction |
| DeepChrome-CNN [11] | × | ✓ | ✓ | Automatic | ✓ | ✓ | ✓ | × | Classification |
| AttentiveChrome[12] | × | ✓ | ✓ | Automatic | ✓ | ✓ | ✓ | ✓ | Classification |
| **DeepDiff** (this study) | ✓ | ✓ | ✓ | Automatic | ✓ | ✓ | ✓ | ✓ | Regression |

Table 2. Comparison of previous studies for the task of quantifying gene expression using histone modification marks (adapted from [11]). The columns indicate (a) whether the it is a differential gene expression or cell type specific gene expression prediction study, (b) whether the study has a unified end-to-end architecture or not (c) if it captures non-linearity among features (d) how has the bin information been incorporated (e) if representation of features is modeled on local and global scales, (f) if combinatorial interactions among histone modifications are modeled, (h) if the model is interpretable, and (g) the output formulation of the study.

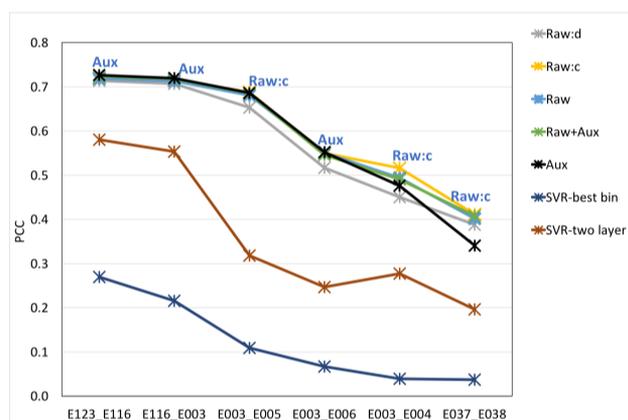

**Fig. 2.** Cell-Specific Auxiliary as $classification$: Pearson correlation (PCC) for DeepDiff main task and multi-tasking with Cell-Specific Auxiliary as classification for six cell-type pairs. The text label for each cell-type pair is the best performing DeepDiff variation.

| Histone Mark | Associated with Regions |
|---|---|
| H3K4me3 | Promoter |
| H3K4me1 | Enhancer |
| H3K36me3 | Transcribed |
| H3K9me3 | Heterochromatin |
| H3K27me3 | Polycomb Repression |

Table 4. Five core histone modifications as defined by [9] with associated regions on the genome.

| REMC Id | Cell type |
|---|---|
| E123 | K562 |
| E116 | GM12878 |
| E003 | H1 Cell Line |
| E004 | H1 BMP4 Derived Mesendoderm Cultured Cells |
| E005 | H1 BMP4 Derived Trophoblast Cultured Cells |
| E006 | H1 Derived Mesenchymal Stem Cells |
| E037 | CD4 Memory Primary Cells |
| E038 | CD4 Naive Primary Cells |
| E007 | H1 Derived Neuronal Progenitor Cultured Cells |

Table 5. The cell-types (and corresponding REMC ID) used in the experiments.

### References

[1] Chao Cheng and Mark Gerstein. Modeling the relative relationship of transcription factor binding and histone modifications to gene expression levels

| Method | Mean | Median |
|---|---|---|
| $Aux$ | 173.20 | 172.58 |
| $Raw+Aux$ | 179.17 | 192.33 |

Table 3. Relative performance with Cell-Specific Auxiliary as $classification$: Mean and Median of the relative performance (%) with respect to Pearson Correlation Coefficient(PCC) when comparing DeepDiff multitasking with classification based Cell-Specific Auxiliary models to one of the best-performing baselines: two-layer SVR across $six$ cell-type pairs.

| Pairs(Tasks) | Cell type A | Cell type B |
|---|---|---|
| 1 | E123 | E003 |
| 2 | E116 | E003 |
| 3 | E123 | E116 |
| 4 | E003 | E005 |
| 5 | E003 | E006 |
| 6 | E006 | E007 |
| 7 | E005 | E006 |
| 8 | E003 | E004 |
| 9 | E004 | E006 |
| 10 | E037 | E038 |

Table 6. The Cell-type pairs we use in our experiments.

in mouse embryonic stem cells. *Nucleic acids research*, 40(2):553–568, 2011.

[2] Chao Cheng, Koon-Kiu Yan, Kevin Y Yip, Joel Rozowsky, Roger Alexander, Chong Shou, Mark Gerstein, et al. A statistical framework for modeling gene expression using chromatin features and application to modencode datasets. *Genome Biol*, 12(2):R15, 2011.

[3] Xianjun Dong, Melissa C Greven, Anshul Kundaje, Sarah Djebali, James B Brown, Chao Cheng, Thomas R Gingeras, Mark Gerstein, Roderic Guigó, Ewan Birney, et al. Modeling gene expression using chromatin features in various cellular contexts. *Genome Biol*, 13(9):R53, 2012.

[4] Raia Hadsell, Sumit Chopra, and Yann LeCun. Dimensionality reduction by learning an invariant mapping. In *Computer vision and pattern recognition, 2006 IEEE computer society conference on*, volume 2, pp. 1735–1742. IEEE, 2006.

[5] Bich Hai Ho, Rania Mohammed Kotb Hassen, and Ngoc Tu Le. Combinatorial roles of dna methylation and histone modifications on gene expression. In *Some Current Advanced Researches on Information and Computer Science in Vietnam*, pp. 123–135. Springer, 2015.

[6] Rosa Karlić, Ho-Ryun Chung, Julia Lasserre, Kristian Vlahoviček, and Martin Vingron. Histone modification levels are predictive for gene expression. *Proceedings of the National Academy of Sciences*, 107(7):2926–2931, 2010.

[7] Diederik Kingma and Jimmy Ba. Adam: A method for stochastic optimization. 2014.

[8] Alex Krizhevsky, Ilya Sutskever, and Geoffrey E Hinton. Imagenet classification with deep convolutional neural networks. In *Advances in neural information processing systems*, pp. 1097–1105, 2012.

[9] Anshul Kundaje, Wouter Meuleman, Jason Ernst, Misha Bilenky, Angela Yen, Alireza Heravi-Moussavi, Pouya Kheradpour, Zhizhuo Zhang, Jianrong Wang, Michael J Ziller, et al. Integrative analysis of 111 reference human epigenomes. *Nature*, 518(7539):317–330, 2015.

[10] Jeffery Li, Travers Ching, Sijia Huang, and Lana X Garmire. Using epigenomics data to predict gene expression in lung cancer. In *BMC bioinformatics*, volume 16, p. S10. BioMed Central, 2015.

[11] Ritambhara Singh, Jack Lanchantin, Gabriel Robins, and Yanjun Qi. Deepchrome: deep-learning for predicting gene expression from histone modifications. *Bioinformatics*, 32(17):i639–i648, 2016.

[12] Ritambhara Singh, Jack Lanchantin, Arshdeep Sekhon, and Yanjun Qi. Attend and predict: Understanding gene regulation by selective attention on chromatin. In I. Guyon, U. V. Luxburg, S. Bengio, H. Wallach, R. Fergus, S. Vishwanathan, and R. Garnett, editors, *Advances in Neural Information Processing Systems 30*, pp. 6785–6795. Curran Associates, Inc., 2017.